\pdfoutput=1

\documentclass[11pt]{article}

\usepackage[preprint]{acl}

\usepackage{times}
\usepackage{latexsym}
\usepackage{tikz}
\usepackage{subcaption}
\usepackage{hhline}
\usepackage{makecell}
\usepackage{amsfonts} 
\usepackage{cuted}

\usepackage[T1]{fontenc}

\usepackage[utf8]{inputenc}
\usepackage{amsmath}
\usepackage{xspace}
\usepackage{colortbl}
\usepackage{rotating}

\usepackage{microtype}

\usepackage{inconsolata}
\usepackage{booktabs} 

\definecolor{skyblue}{RGB}{135,206,235}

\usepackage{graphicx}

\title{
Model Unlearning Objectives Vary for Distinct Language Functions}

\author{%
Berk Atil\textsuperscript{$1$} \enspace Vipul Gupta \textsuperscript{$2$} \enspace Rebecca J. Passonneau\textsuperscript{$1$} 
\\
{\textsuperscript{$1$} {Pennsylvania State University} \textsuperscript{$2$} {Scale AI}}\quad \\
\quad \\
\normalsize{\tt bka5352@psu.edu}\\
}

\begin{document}
\maketitle
\begin{abstract}
Large language models (LLMs) learn undesirable properties during pretraining, including dangerous knowledge and toxic text generation. Just as post-training uses different objectives to shape different behaviors, we argue that unlearning methods should be designed for the language function at issue. To study this, we consider two mechanistically distinct unlearning goals, dangerous-knowledge unlearning and toxicity unlearning.
For dangerous knowledge, we introduce a cosine-based, meta-learned variant of RMU. For toxicity, we propose a multi-layer objective based on layer-specific probe directions.
Across four open-source 7-8B models, our methods 
achieve strong results, based on distinct training objectives for the two types of unlearning.
Overall, our results suggest that unlearning should be studied as a family of problems, analogous to the multiple types of LLM post-training.
\end{abstract}
    
\section{Introduction}

Language serves a wide range of functions in human society: it is used not only to convey information, but also to coordinate action, manage social relationships, and express attitudes, intentions, and norms.
Large language models (LLMs) acquire great linguistic fluency through large-scale pretraining, and post-training methods are then used to shape behaviors such as helpfulness, instruction-following, and safer response behavior \cite{ouyang2022instructgpt,rafailov2023dpo,du2025posttraining}. However, these methods 
do not cover the full range of communicative abilities and norms that human language use fulfills.
Models learn hazardous knowledge and socially harmful behaviors, such as toxic language generation, from their training data \cite{brown2020gpt3,gehman2020realtoxicityprompts,li2024wmdp}. 
This has motivated growing interest in \emph{unlearning}: post hoc interventions that aim to remove specific knowledge, capabilities, or behaviors from already-trained models 
through finetuning 
\cite{cao2015towards,bourtoule2021machine,liu2024rethinking,maini2024tofu}. 

Our central claim is that, similar to post-training itself, \emph{unlearning is goal-dependent}. Modern LLM pipelines 
have distinct post-training procedures for 
a range of distinct
functions:
instruction following, preference alignment to human values, refusal behavior, and style control require different objectives 
\cite{ouyang2022instructgpt,rafailov2023dpo,du2025posttraining}. Mechanistic work further suggests that some of these properties are altered more substantially by post-training than others. In particular, \citet{du2025posttraining} show that factual knowledge storage locations remain largely stable across base and post-trained models, and that a truthfulness direction is also highly similar across the two, whereas refusal directions change substantially after SFT and instruction tuning. We argue that unlearning should take such redmechanistic observations into account. Removing dangerous knowledge is not the same problem as removing socially undesirable behavior, because 
these functions are represented differently inside the model.

We study two unlearning problems that mechanistic evidence suggests could be different: removal of \emph{dangerous knowledge} versus \emph{toxic language}. Dangerous knowledge concerns the model's ability to 
access factual or procedural information, as in biosecurity-oriented settings such as WMDP \cite{li2024wmdp}. Toxicity, by contrast, concerns the tendency to generate abusive or harmful language 
\cite{gehman2020realtoxicityprompts,hartvigsen2022toxigen}. 
Existing work \cite{kadhe2024split} has 
treated both as instances of the same unlearning problem, but our results suggest otherwise, which is supported by recent mechanistic studies. 
For factual knowledge, prior work points to a relatively structured retrieval mechanism \cite{meng2022rome}. Knowledge relevant to a statement is concentrated at subject, object, and last-token positions, with subject information strongest in earlier layers, object information in early-to-middle layers, and the last token becoming especially important in middle-to-late layers \cite{meng2022rome,geva2023dissecting,du2025posttraining}.  \citet{du2025posttraining} further show that post-training largely preserves these knowledge-storage locations. At the same time, current hazardous-knowledge unlearning methods remain limited: they can be shallow or recoverable, suggesting that simply steering away from undesirable behavior might not be enough \cite{hu2024jogging,deeb2024remove,dang2024effects,dang2025robustness}. 

The mechanisms 
underlying toxicity in LLMs differ from those
for factual knowledge. 
\citet{lee2024dpo} train a linear toxicity probe on averaged final-layer representations and identify value vectors aligned with a toxicity direction. They show that toxicity is largely elicited in later MLP layers, and that subtracting those vectors can reduce toxic outputs. Crucially, after DPO, the toxic vectors largely remain; instead, small accumulated changes in the model alter activations so that the model bypasses toxicity-promoting regions rather than erasing the underlying capability \cite{lee2024dpo}. 
These findings suggest \textbf{complementary roles for alignment and unlearning}: alignment can reduce toxic behavior, while unlearning can 
weaken the underlying toxicity-relevant directions. 

Motivated by this difference, we argue that effective unlearning requires a better understanding of \emph{what is being forgotten}. At a high level, unlearning methods aim to reduce undesirable behavior while preserving desirable behavior, which is reflected in the standard combination of a forget objective and a retain objective used across prior work \cite{li2024wmdp,huu2024effects,zhang2024npo}. 
We claim, however, that the formulation of
the forget objective should depend on the unlearning goal. 
We introduce a new unlearning method for dangerous knowledge 
that replaces RMU's L2 objective \cite{li2024wmdp} with a cosine objective, and adaptively learns the retain-forget tradeoff using reinforcement learning. 
For toxicity unlearning,
we propose a forget loss that 
targets 
toxicity-relevant signals across multiple layers. 
For assessment, we introduce a unified evaluation metric that summarizes the tradeoff between forgetting and retention. 


\section{Related Work}
\vspace{-0.1cm}
In this section, we review work 
to understand how LLMS store undesirable knowledge and toxicity. 

\subsection{LLM Unlearning}

Unlearning was originally framed as removing the influence of specific training data without full retraining \cite{cao2015towards,bourtoule2021machine}. In LLMs, exact retraining-based guarantees are usually infeasible, so recent work instead relies on 
benchmark-specific forgetting metrics \cite{liu2024rethinking}. Benchmarks such as TOFU \cite{maini2024tofu} and WMDP \cite{li2024wmdp} have become standard datasets for fictitious-profile forgetting and hazardous-knowledge unlearning, respectively. Existing methods include gradient-based forget objectives, preference-based methods such as NPO \cite{zhang2024npo}, and representation-level steering approaches such as RMU \cite{li2024wmdp} or Spunge \cite{kadhe2024split}.

\subsection{Unlearning Dangerous Knowledge}

A major line of work studies unlearning of dangerous factual or procedural knowledge, especially through WMDP-style evaluations \cite{li2024wmdp}. However, 
unlearned knowledge can often be recovered through targeted relearning, substantial information may remain in model weights, and steering-based methods can reduce robustness or induce nonsensical behavior rather than cleanly removing the target capability \cite{hu2024jogging,deeb2024remove,dang2024effects,dang2025robustness}. Our method 
aims to provide a more principled objective for hazardous-knowledge unlearning.

Mechanistic work helps clarify 
the problem of unlearning dangerous knowledge.
Factual recall depends strongly on subject, object, and last-token positions across different layer ranges \cite{meng2022rome,geva2023dissecting,du2025posttraining}. \citet{du2025posttraining} further shows that post-training largely preserves these knowledge-storage locations, suggesting that factual competence is anchored in relatively stable internal structure. 
\citet{zou2023representation} found that representation-level interventions offer more control on safety-relevant capabilities. 
This motivates our representation-level approach that 
aims to weaken dangerous factual competence while preserving general utility, by directly targeting representations that support factual recall.

\subsection{Toxicity Behaviour}
\label{sec:rel_work_tox}
A separate line of work studies harmful language generation, including toxicity and abusive behavior \cite{gehman2020realtoxicityprompts,hartvigsen2022toxigen}. Mechanistic evidence suggests that toxicity is represented differently from factual knowledge. \citet{lee2024dpo} show that a toxicity direction can be extracted from final-layer hidden states and that later-layer value vectors aligned with this direction can modulate toxic outputs. Yet apparently, DPO does not remove these toxic vectors; instead, it introduces distributed offsets that bypass toxicity-eliciting regions \cite{lee2024dpo}. More generally, this suggests that alignment may reduce toxic behavior without fully removing the underlying toxic capability. 
\citet{du2025posttraining} similarly show that refusal directions shift substantially across post-training, unlike truthfulness directions, reinforcing the view that safety behaviors are more post-training-dependent and less structurally stable than factual knowledge. 

When we tried an RMU-like approach for toxicity, the results were poor. Toxicity is distributed across layers, this %
motivates our multi-layer toxicity unlearning objective. 

\begin{equation}    
\small
L = L_{\text{F}} + \alpha \cdot L_{\text{R}}
\label{eq:total_loss}
\end{equation}

\begin{equation}   
\small
L_{\text{F}} = \mathbb{E}_{x \sim D_{\text{F}}}\left[\sum_{t \in x_f} \|M_{\text{updated}}(t) - c \cdot \mathbf{u}\|_2^2 \right]
\label{eq:forget_loss}
\end{equation}

\begin{equation}
\small
L_{\text{R}} = \mathbb{E}_{x\sim D_{\text{R}}}\left[\sum_{t \in x_r} \|M_{\text{updated}}(t) - M_{\text{frozen}}(t)\|_2^2 \right]
\label{eq:retain_loss}
\end{equation}

\section{Baselines For Unlearning}

In this section, we review the main baseline approaches that we build upon. 

\subsection{RMU}

Representation Misdirection for Unlearning (RMU) is a fine-tuning approach aimed at selectively removing unwanted knowledge from LLMs \cite{li2024wmdp}. RMU operates on two datasets:  \emph{forget data}, containing the target knowledge or behavior to be unlearned, and \emph{retain data}, containing general examples used to preserve the model’s desirable capabilities. It pushes the model’s representations on forget data toward randomly initialized vectors, while also encouraging representations on retain data to remain close to those of the original frozen model. 
The overall objective is shown in Equation \ref{eq:total_loss}. The forget loss is shown in Equation \ref{eq:forget_loss} where $\mathbf{u}$ is a random unit vector and $c$ is a scaling factor. Similarly, $L_{\text{retain}}$ refers to the retain loss and is taken as an expectation over the entire retain dataset, as shown in Equation \ref{eq:retain_loss}.

\subsection{AdapRMU}

\citet{huu2024effects}, focus on the scaling coefficient $c$ in Equation \ref{eq:forget_loss}. While the direction $\mathbf{u}$ is fixed before unlearning, $c$ determines the magnitude of the representation shift. AdapRMU adaptively adjusts $c$ based on the norm of the forget representation. 

\section{Methodology}
\label{sec:method}
In this section, we present our modifications to RMU. We make two changes for dangerous-knowledge unlearning and then introduce a separate objective for toxicity. 
We also present our metric to reflect the tradeoff between unlearning and general capability.

\begin{equation}
    \scriptsize
    \mathcal{L}_{\text{F}} = \mathbb{E}_{x_f \sim D_{\text{F}}}\left[\sum_{t \in x_f}\left(1 - \frac{M_{\text{updated}}(t) \cdot (c \cdot \mathbf{u})}{\|M_{\text{updated}}(t)\|_2 \|c \cdot \mathbf{u}\|_2}\right)\right]
\label{eq:cosine_forget_loss}
\end{equation}

\begin{equation}
\tiny
\mathcal{L}_{\text{R}} = \mathbb{E}_{x_r \sim D_{\text{R}}}\left[\sum_{t \in x_r}\left(1 - \frac{M_{\text{updated}}(t) \cdot M_{\text{frozen}}(t)}{\|M_{\text{updated}}(t)\|_2 \|M_{\text{frozen}}(t)\|_2}\right)\right]
\label{eq:cosine_retain_loss}
\end{equation}

\subsection{Cosine Loss instead of L2 Loss}

We replace the original L2 losses in Equations \ref{eq:forget_loss} and \ref{eq:retain_loss} with cosine-based losses
shown in Equations~\ref{eq:cosine_forget_loss} and~\ref{eq:cosine_retain_loss}. 
Our hypothesis is that the key quantity for both forgetting and retention is representational \emph{direction} rather than exact Euclidean position. For forgetting, cosine loss should work better because it directly optimizes alignment between the updated representation and the target direction, instead of also penalizing norm differences. For retention, cosine loss better preserves the frozen model's representational orientation on retain data while tolerating harmless changes in magnitude induced by fine-tuning. In this sense, cosine-based losses are a better match to our intended geometry: forgetting should steer representations toward or away from a direction, and retention should preserve directional structure rather than exact scale. Prior work similarly notes that L2 distance can be mismatched when the task is fundamentally angular, while cosine-based objectives can be more robust to norm variation \cite{xu2018robust}. 







\subsection{Meta Learning the Coefficient $\alpha$} 
In Equation \ref{eq:total_loss}, the coefficient $\alpha$ balances the forget and retain objectives. In RMU and AdapRMU, $\alpha$ is fixed throughout training. We found this to be a limitation, because we observed the best value of $\alpha$ to vary substantially across models and datasets. To address this, we treat $\alpha$ as a learnable parameter and update it during fine-tuning using REINFORCE \cite{williams1992simple}. At each step, the policy update adjusts $\alpha$ toward values that improve the overall objective, allowing the model to adaptively balance forgetting and retention. This removes the need for extensive manual tuning and leads to more consistent performance across models.

\begin{figure}[t!]
    \centering
    \includegraphics[width=0.49\textwidth]{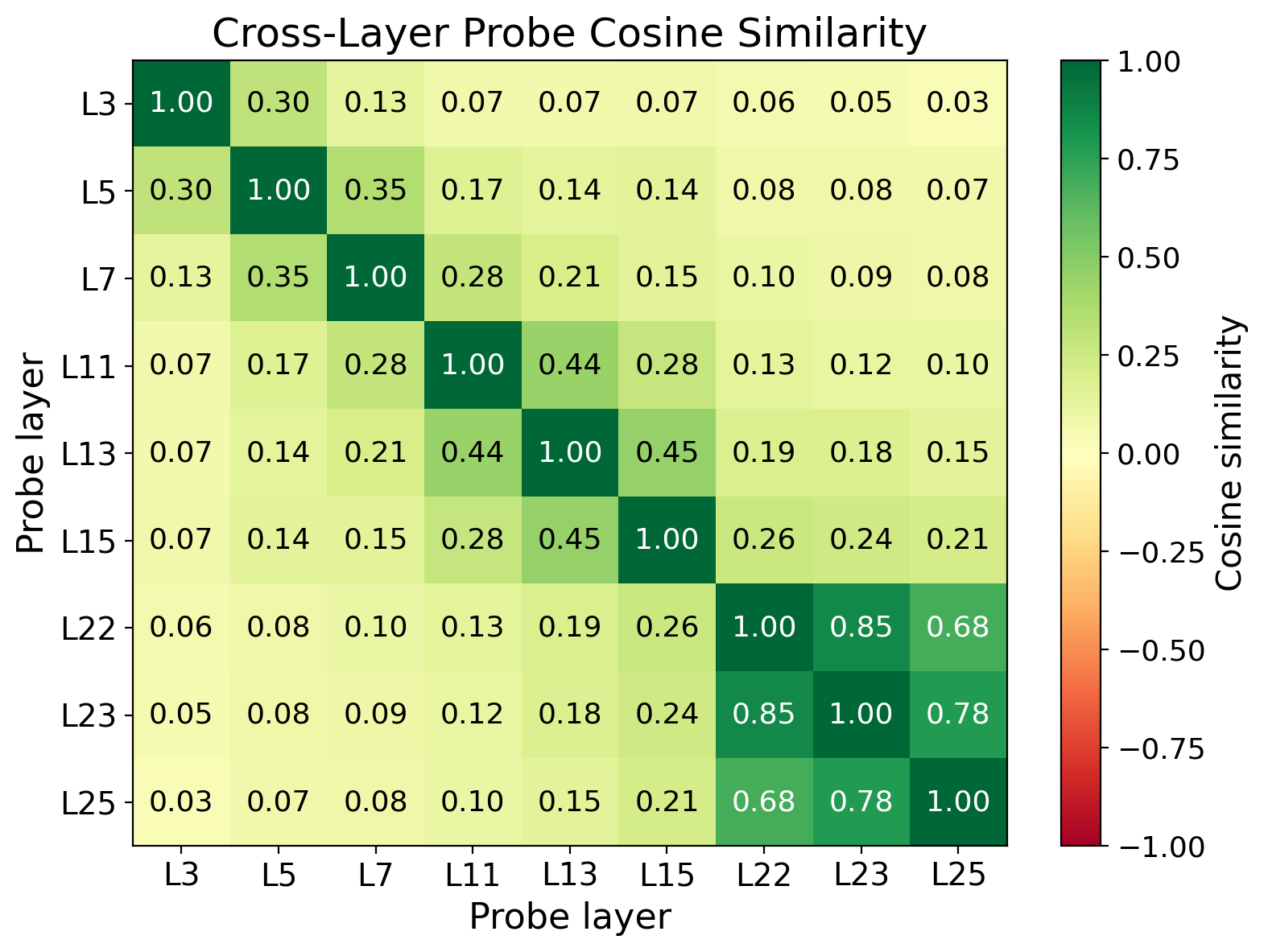}
    \caption{\small Cosine similarities between logistic-regression probes trained at different layers of Llama3.1-8B.}
    \label{fig:probe_similarity}
\end{figure}

\begin{figure}[t!]
    \centering
    \includegraphics[width=0.49\textwidth]{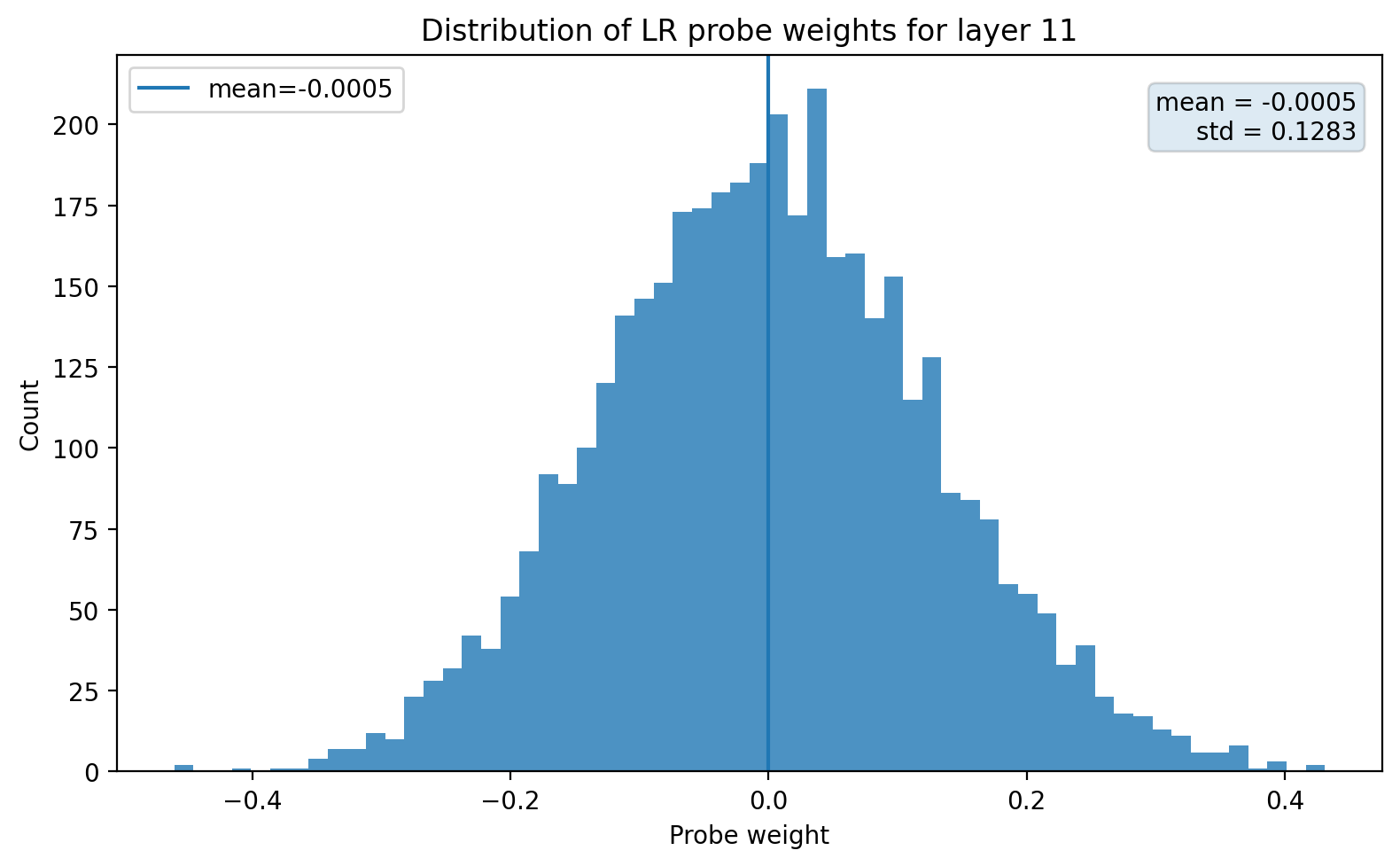}
    \caption{\small Weight distribution of the logistic-regression probe trained on layer 11 of Llama3.1-8B for toxicity classification.}
    \label{fig:probe_weight_analysis}
\end{figure}

\begin{figure*}[t]
\centering

\begin{subfigure}[b]{0.45\textwidth}
    \centering
    \includegraphics[width=\textwidth]{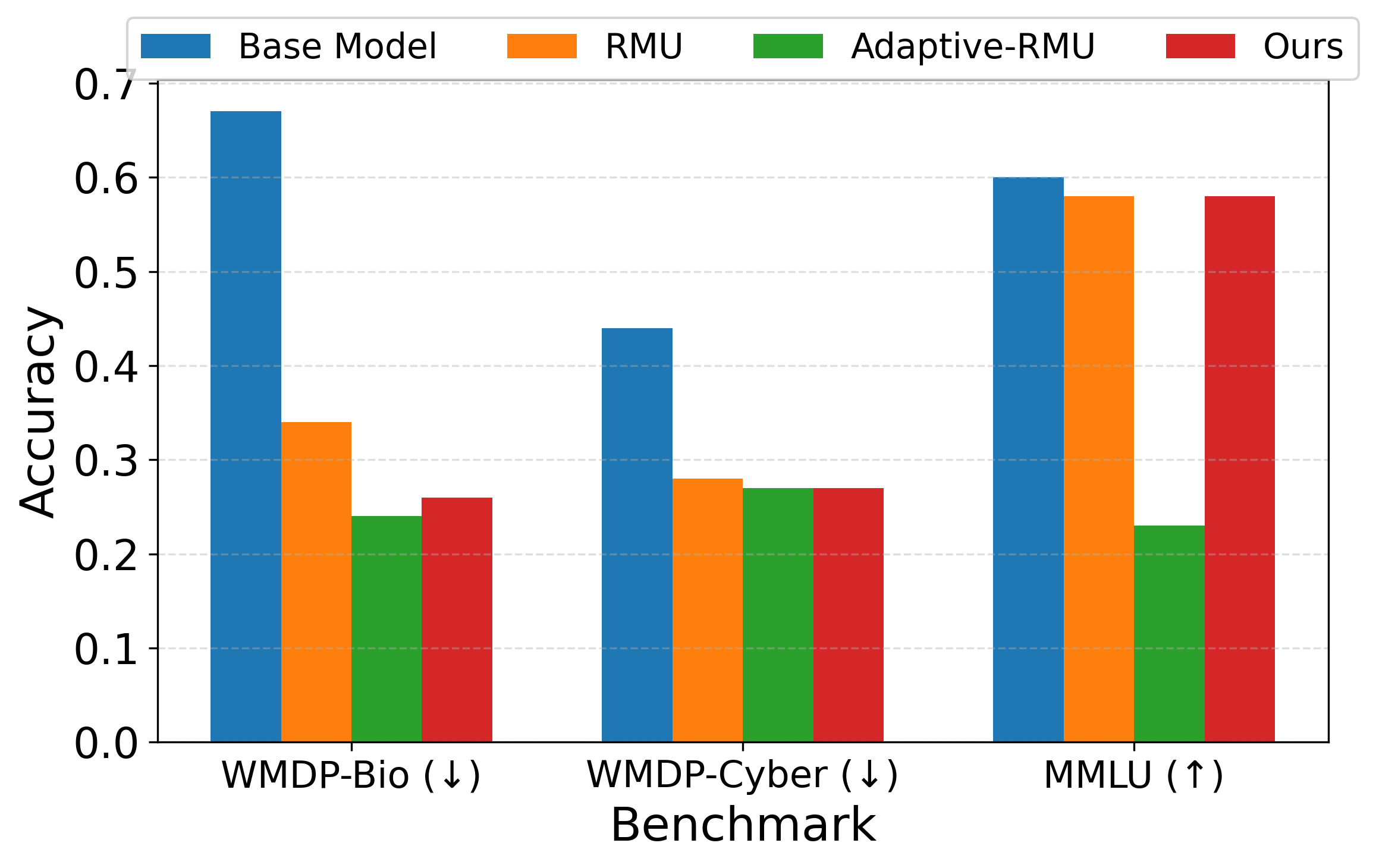}
    \caption{Mistral-7B}
\end{subfigure}
\hfill
\begin{subfigure}[b]{0.45\textwidth}
    \centering
    \includegraphics[width=\textwidth]{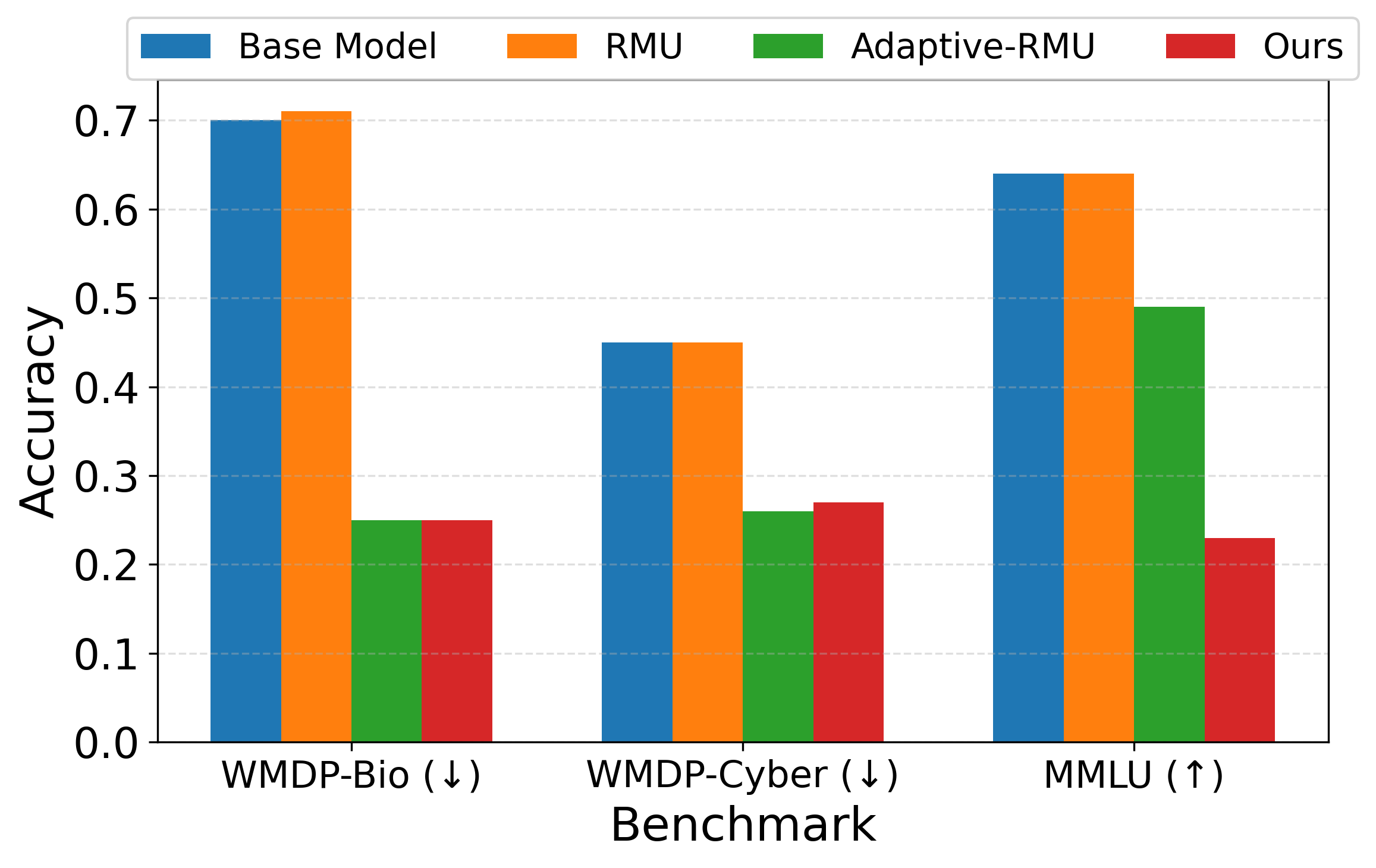}
    \caption{Llama3.1-8B}
\end{subfigure}

\vspace{0.5cm}

\begin{subfigure}[b]{0.45\textwidth}
    \centering
    \includegraphics[width=\textwidth]{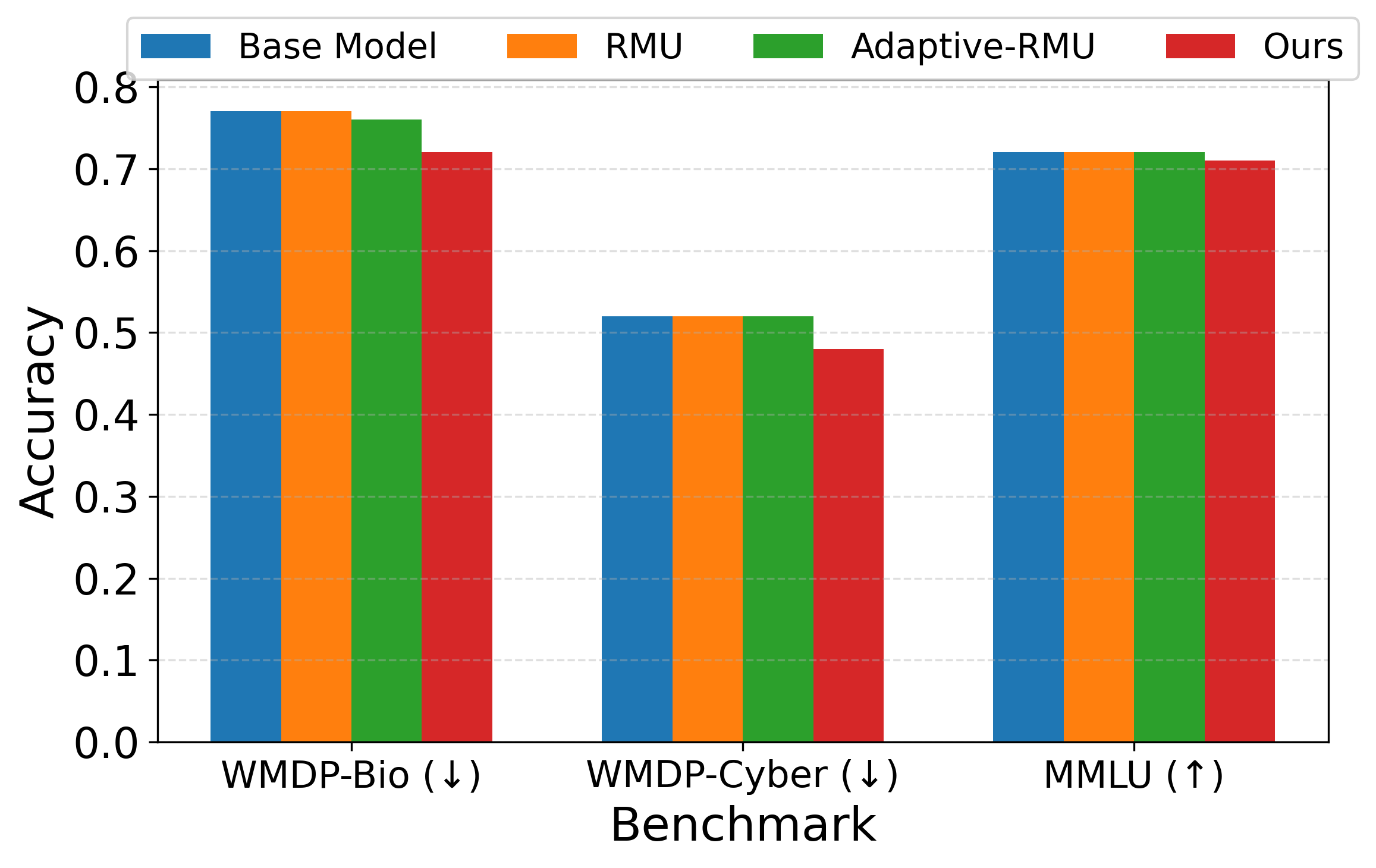}
    \caption{Qwen2.5-7B}
\end{subfigure}
\begin{subfigure}[b]{0.45\textwidth}
    \centering
    \includegraphics[width=\textwidth]{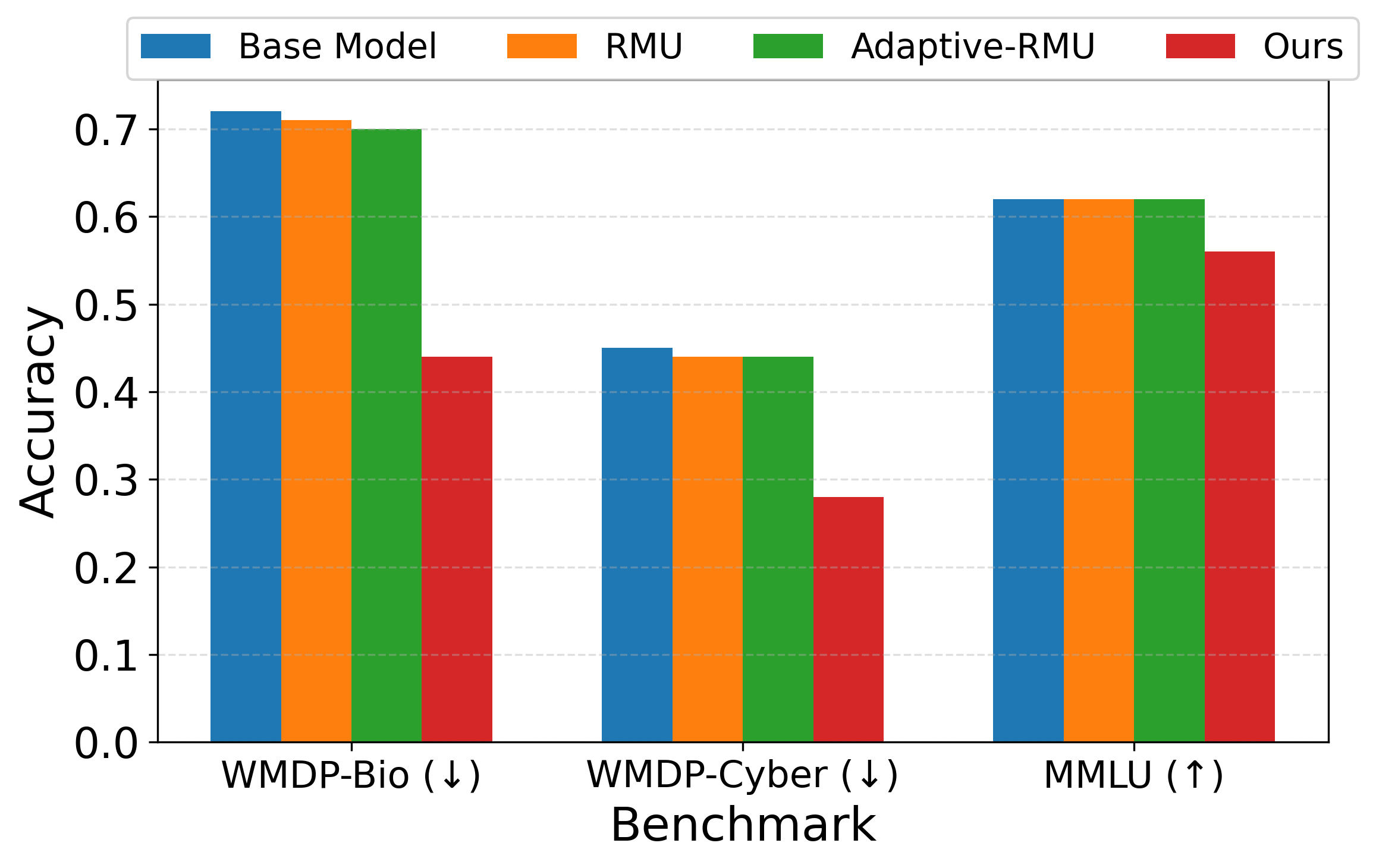}
    \caption{Olmo3-7B}
\end{subfigure}

\caption{Dangerous knowledge unlearning results for the four models.}
\label{fig:danger_un_results}
\end{figure*}

In more detail, REINFORCE evaluates how the $\alpha$ value affects the total loss, then adjusts $\alpha$ in the direction that lowers the loss most effectively. 
In this way, the model learns the best possible balance between forgetting unwanted information and preserving useful knowledge. The result is an $\alpha$ value that adapts automatically to different models, thus yielding more consistent improvement. 

\begin{figure*}[t]
\centering

\begin{subfigure}[b]{0.45\textwidth}
    \centering
    \includegraphics[width=\textwidth]{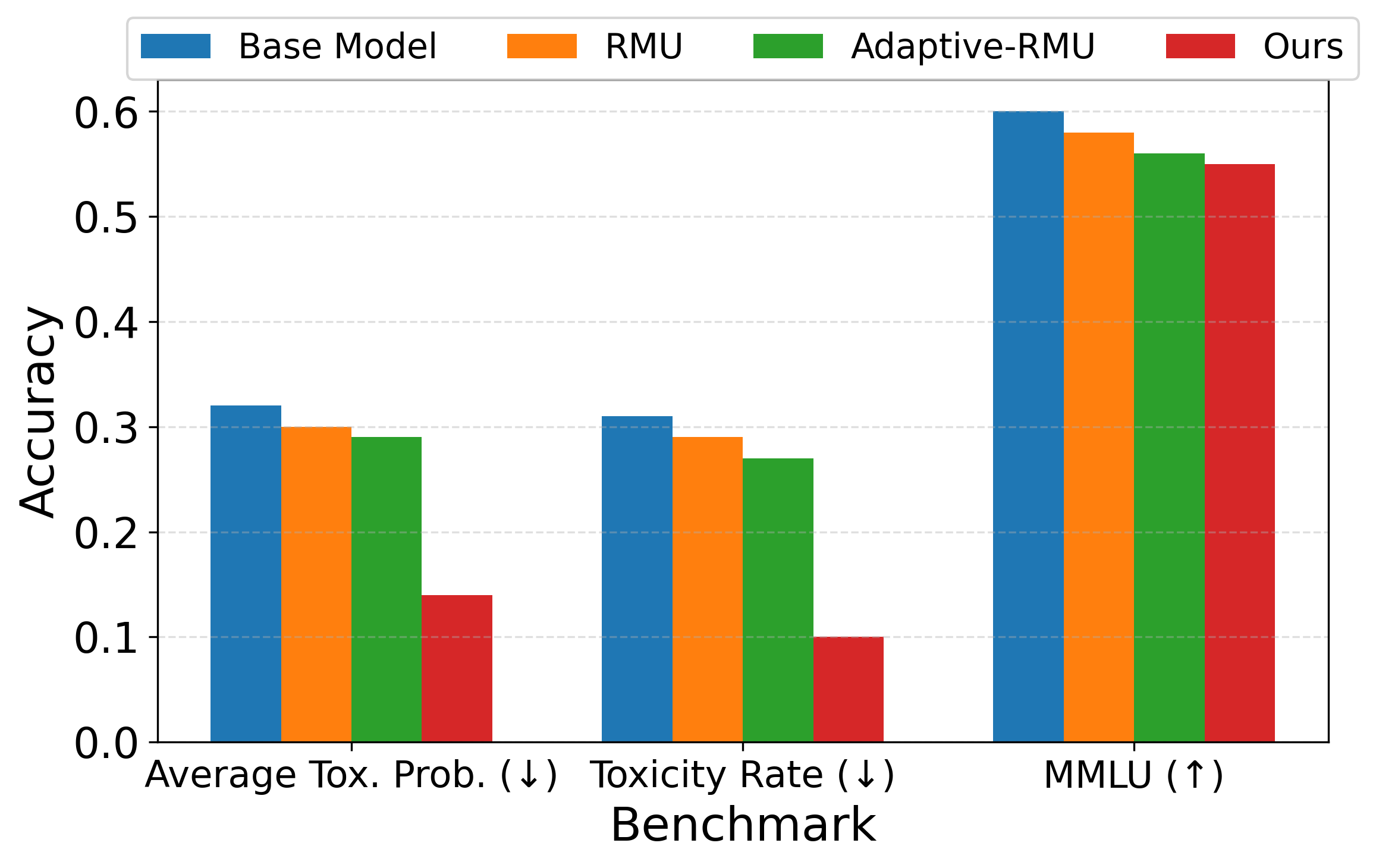}
    \caption{Mistral-7B}
\end{subfigure}
\hfill
\begin{subfigure}[b]{0.45\textwidth}
    \centering
    \includegraphics[width=\textwidth]{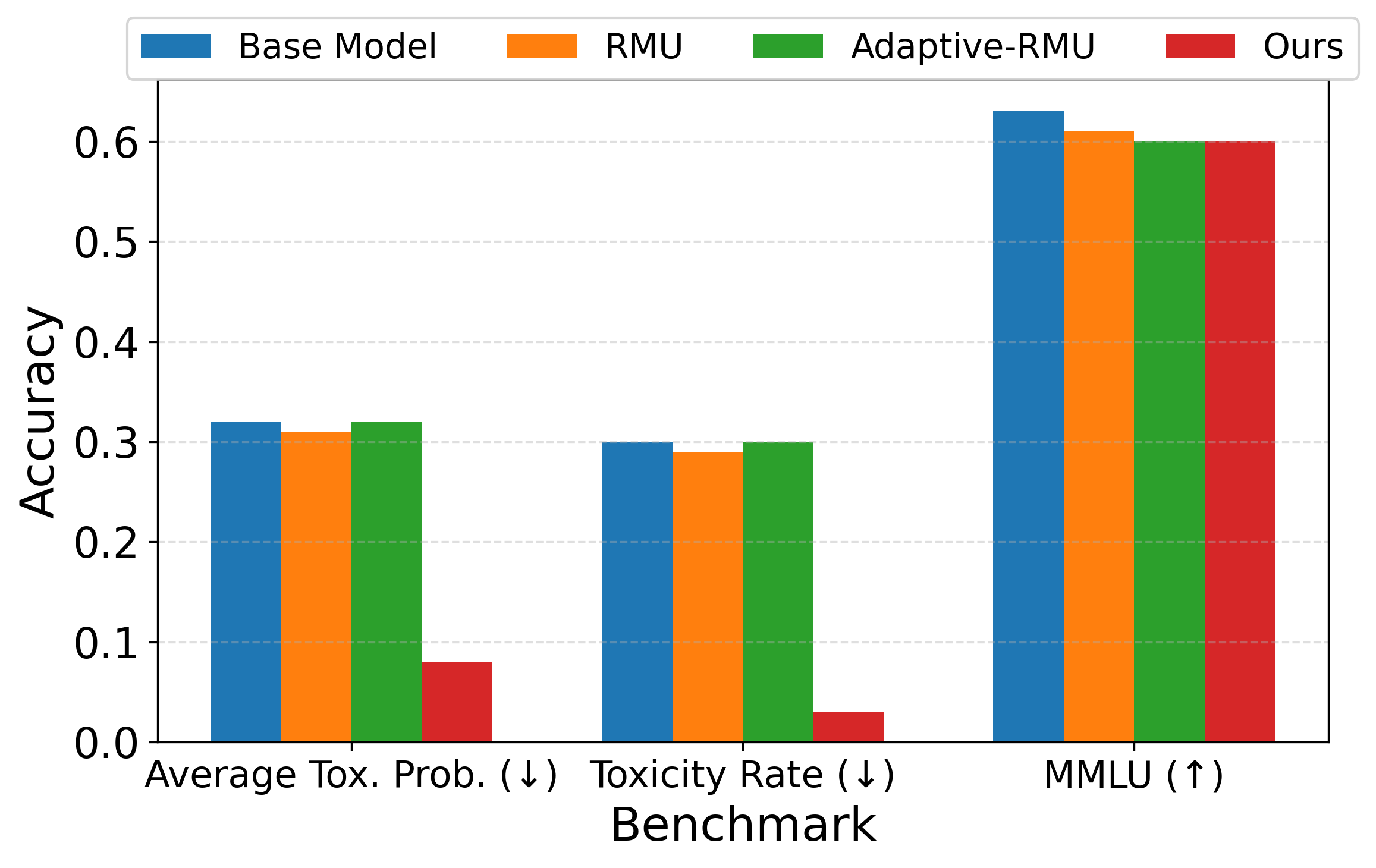}
    \caption{Llama3.1-8B}
\end{subfigure}

\vspace{0.5cm}

\begin{subfigure}[b]{0.45\textwidth}
    \centering
    \includegraphics[width=\textwidth]{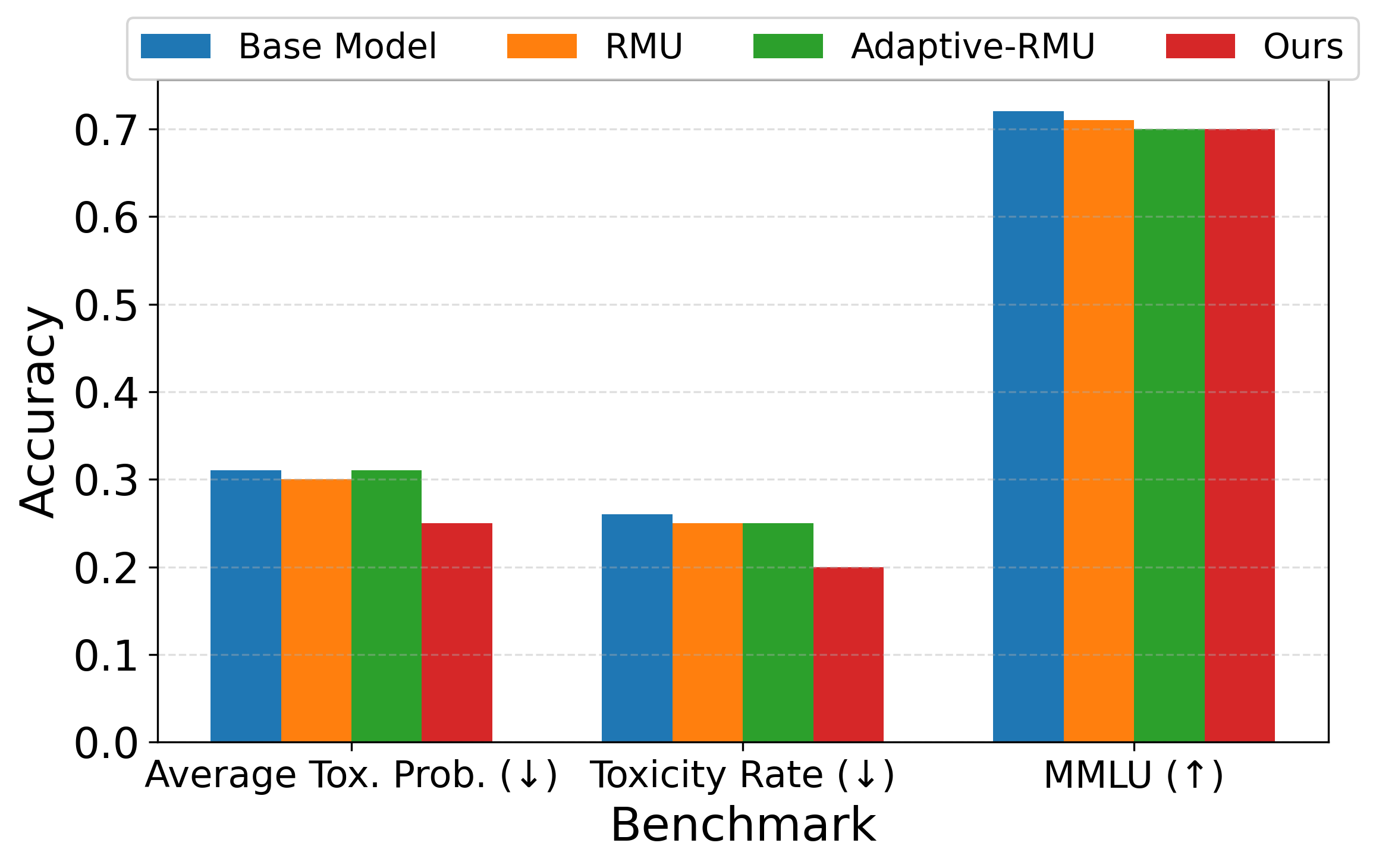}
    \caption{Qwen2.5-7B}
\end{subfigure}
\hfill
\begin{subfigure}[b]{0.45\textwidth}
    \centering
    \includegraphics[width=\textwidth]{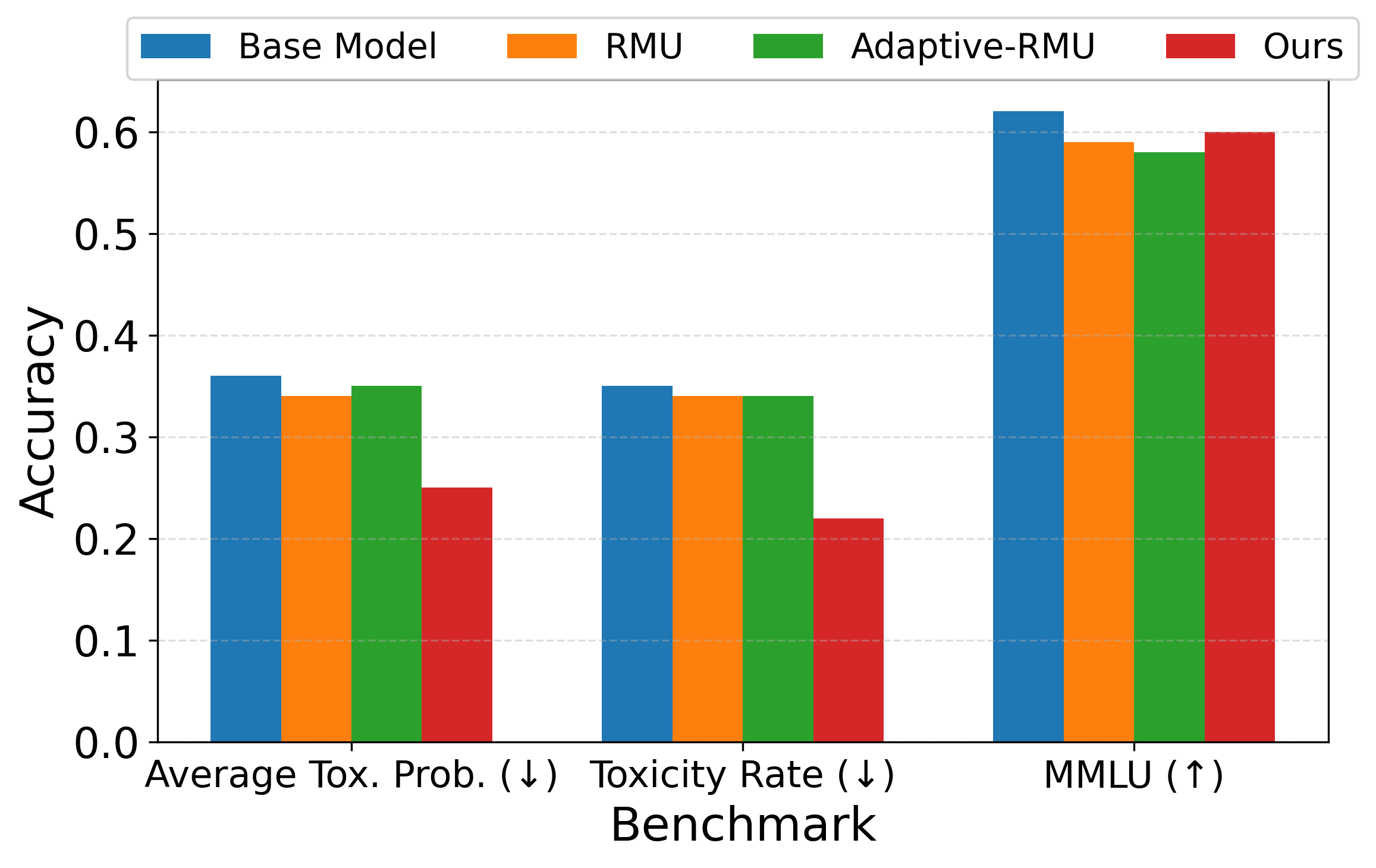}
    \caption{Olmo3-7B}
\end{subfigure}

\caption{Toxicity unlearning results for the four models.}
\label{fig:tox_unlearn_results}
\end{figure*}



\subsection{Toxicity Unlearning}
We initially attempted to use the same loss function for toxicity, but we found 
that simply reusing the same objective designed for hazardous knowledge is ineffective for toxicity (yielding only a 2-3\% decrease in toxicity compared to 6-24\% we find). 

Building on the mechanistic findings in Section \ref{sec:rel_work_tox}, we directly target the empirical toxicity direction in representation space.
We operationalize this by training L2-regularized logistic regression probes on last-token hidden representations from a frozen reference model at multiple layers, obtaining a toxicity direction $LR^i$ for each target layer $i \in I$. We use last-token representations rather than mean pooling because averaging suppresses useful information, which is borne out in our experiments showing that last-token probes consistently achieve higher AUC. We also find that cross-layer cosine similarities between probe directions are often low, indicating that toxicity rotates substantially across depth. This suggests that a single shared direction is not sufficient to characterize toxicity across the network. We therefore train layer-specific probes and intervene at multiple layers at dispersed depths. Below, we report an ablation study for different combinations of layers that verify the utility of both multiple layers, and dispersed depths.


\begin{equation}
\small
\mathcal{L}_{\text{F}} = \mathbb{E}_{x_t \sim D_{\text{toxic}}}\left[\sum_{i \in I}\left( M^i_{\text{updated}}(x_t[-1]) \cdot LR^i \right)^2\right]
\label{eq:toxicity_loss_new}
\end{equation}

Our forget loss, shown in Equation \ref{eq:toxicity_loss_new}, encourages the updated model’s representations of toxic inputs to become orthogonal to the toxicity direction at each layer. 
Squaring the dot product is important for three reasons. First, it ensures a non-negative loss that is minimized when the representation is orthogonal to the toxicity direction. Second, it makes the gradient symmetric around zero. Third, it gives larger gradients to representations that are strongly aligned with the toxicity direction. Averaging across layers in $I$ ensures that the toxic subspace is suppressed at every targeted depth, reducing the chance that toxicity suppressed in one layer re-emerges later in the network.

To preserve general language-modeling ability, we use the same retain loss as in dangerous-knowledge unlearning. The full objective is still given by Equation \ref{eq:total_loss}, with $\alpha$ adapted using the meta-learning procedure described above.

\subsection{S-unlearning:
a New Unlearning Metric}

To summarize the tradeoff between forgetting and retained utility, we introduce \textbf{S-unlearning}. Let $U \in [0,1]$ denote unlearning performance and $R \in [0,1]$ retained utility. For dangerous knowledge, lower WMDP accuracy indicates better unlearning; for toxicity, lower toxicity indicates better unlearning. Because random guessing on our multiple-choice utility benchmark yields $R_0 = 0.25$, we first chance-correct utility as
\[
\bar{R} = \left[\frac{R-R_0}{1-R_0}\right]_0^1.
\]
For dangerous knowledge, we similarly chance-correct unlearning accuracy, while for toxicity we define $\bar{U}=1-U$ so that higher is better. Our final score is
\[
\text{S-Unlearning} = \bar{U}\cdot\bar{R}.
\]
This score lies in $[0,1]$, is maximized when unlearning is strong and retained utility is high, and becomes zero whenever retained utility is at or below chance. Geometrically, it is the area of the rectangle from the origin to $(\bar{U},\bar{R})$ in normalized unlearning-utility space, i.e., the single-point two-dimensional special case of the hypervolume indicator \cite{auger2012hypervolume}. 


\begin{table*}[t!]
\begin{center}
\small
\begin{tabular}{l|r|r}\hline 
 
\multicolumn{1}{c|}{\textbf{Unlearning Method}} & 
\multicolumn{1}{c|}{\textbf{S-unlearning WMDP}} & 
\multicolumn{1}{c}{\textbf{S-unlearning Toxicity}}   \\\hline
\multicolumn{3}{c}{\textbf{Llama3.1-8B}} \\\hline
Base  & 0.29 & 0.34 \\ 
RMU  & 0.29 & 0.33\\ 
Adaptive RMU  & 0.32 & \textbf{0.32}\\ 
Ours  & 0.00 & \textbf{0.43} \\\hline
\multicolumn{3}{c}{\textbf{Mistral-7B}} \\\hline
Base  & 0.28 & 0.32 \\ 
RMU  & 0.40 & 0.31\\ 
Adaptive RMU  & 0.00 & 0.29 \\ 
Ours  & \textbf{0.43} & \textbf{0.34} \\ \hline
\multicolumn{3}{c}{\textbf{Qwen2.5-7B}} \\\hline
Base  & 0.30 & 0.43 \\ 
RMU  & 0.30 & 0.43 \\ 
Adaptive RMU  & 0.30 &  0.41 \\ 
Ours  & \textbf{0.33}& \textbf{0.45}\\ \hline
\multicolumn{3}{c}{\textbf{Olmo3-7B}} \\\hline
Base  & 0.27 & 0.32 \\ 
RMU  & 0.28 & 0.30 \\ 
Adaptive RMU  & 0.28 & 0.29 \\ 
Ours  & \textbf{0.35} & \textbf{0.35} \\ 
\hline
\end{tabular}
\end{center}
\caption{\small S-unlearning scores that combine the tradeoff between unlearning and general capability for both unlearning types.}
\label{tab:s_unlearning_resuts} 
\end{table*}


\section{Experimental Setup}
\vspace{-0.1cm}
Here we describe the datasets, models, and evaluation metrics used for both unlearning settings.

For dangerous-knowledge unlearning, we follow 
\cite{li2024wmdp}: we use the biosecurity and cybersecurity forget datasets released with WMDP, evaluate forgetting on WMDP, and use WikiText as retain data. We measure retained general capabilities on MMLU \cite{hendrycks2021measuringmassivemultitasklanguage}.


\begin{figure*}
\begin{subfigure}[b]{0.45\textwidth}
    \centering
    \includegraphics[width=\textwidth]{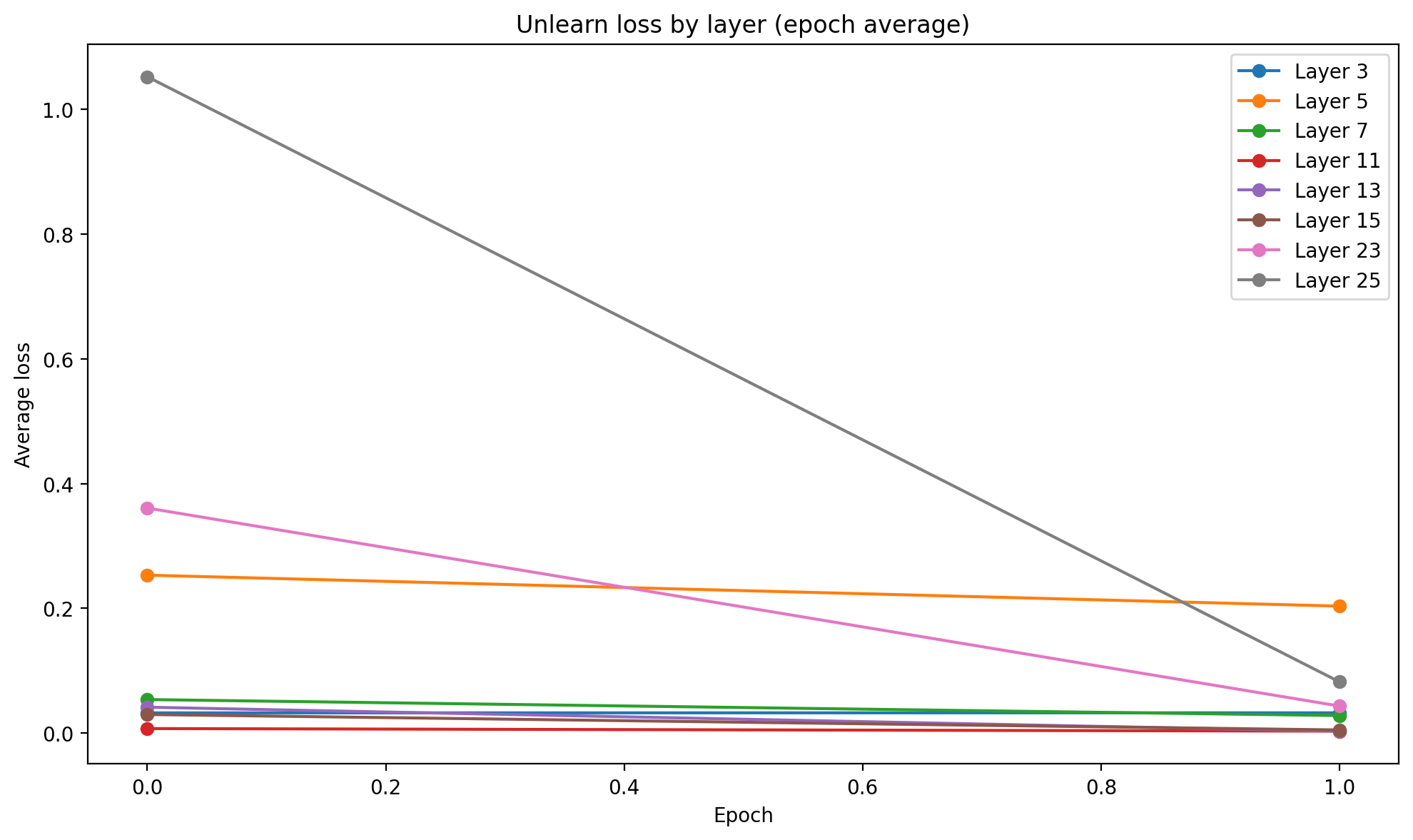}
    \caption{Mistral-7B}
\end{subfigure}
\hfill
\begin{subfigure}[b]{0.45\textwidth}
    \centering
    \includegraphics[width=\textwidth]{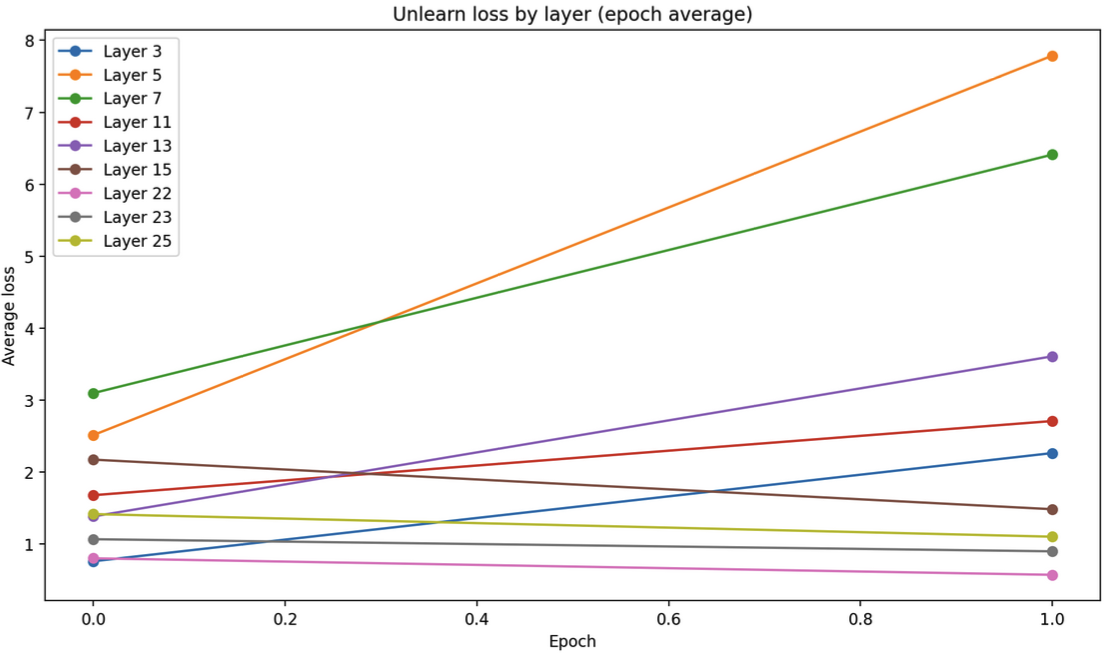}
    \caption{Llama3.1-8B}
\end{subfigure}

\vspace{0.2cm}

\begin{subfigure}[b]{0.45\textwidth}
    \centering
    \includegraphics[width=\textwidth]{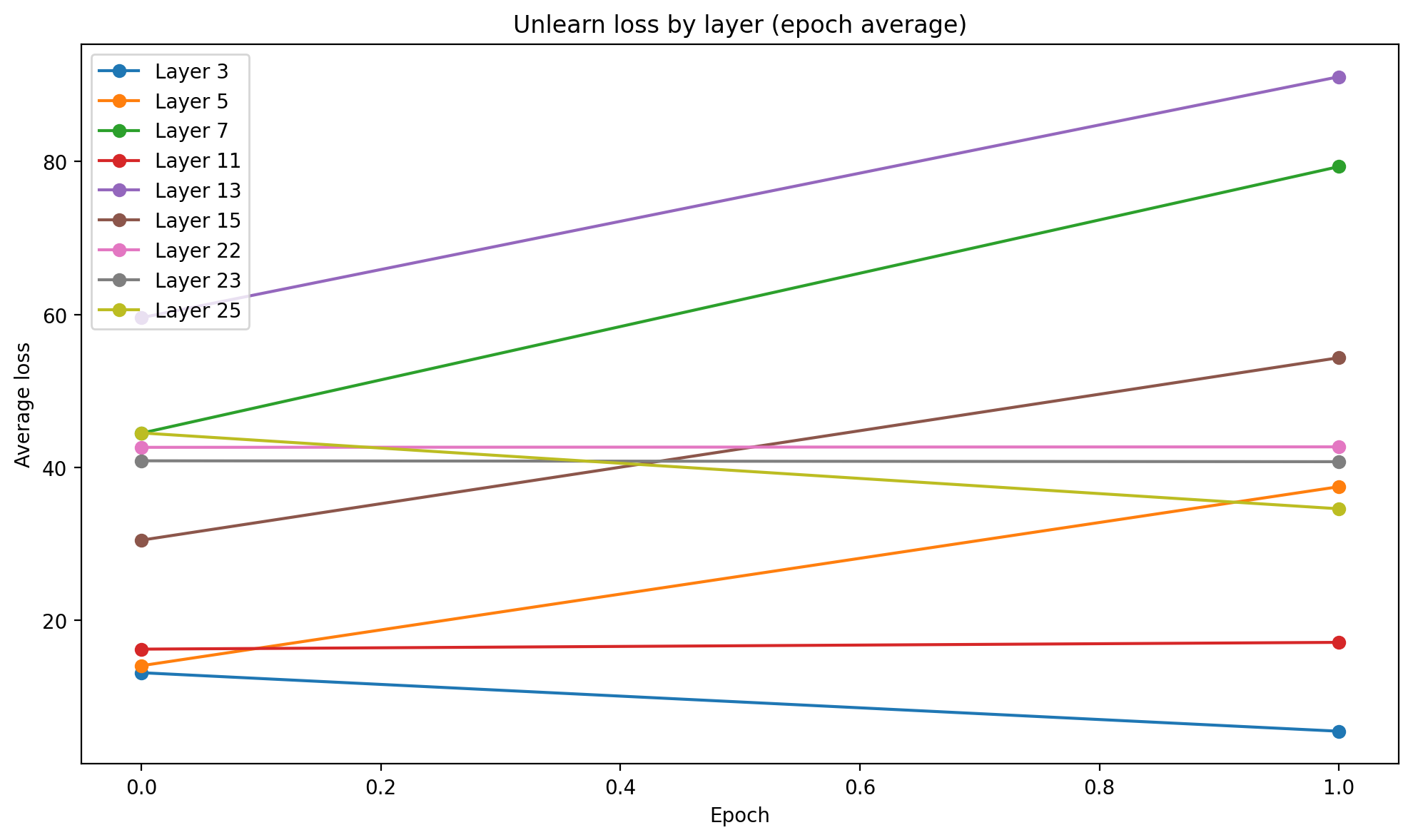}
    \caption{Qwen2.5-7B}
\end{subfigure}
\hfill
\begin{subfigure}[b]{0.45\textwidth}
    \centering
    \includegraphics[width=\textwidth]{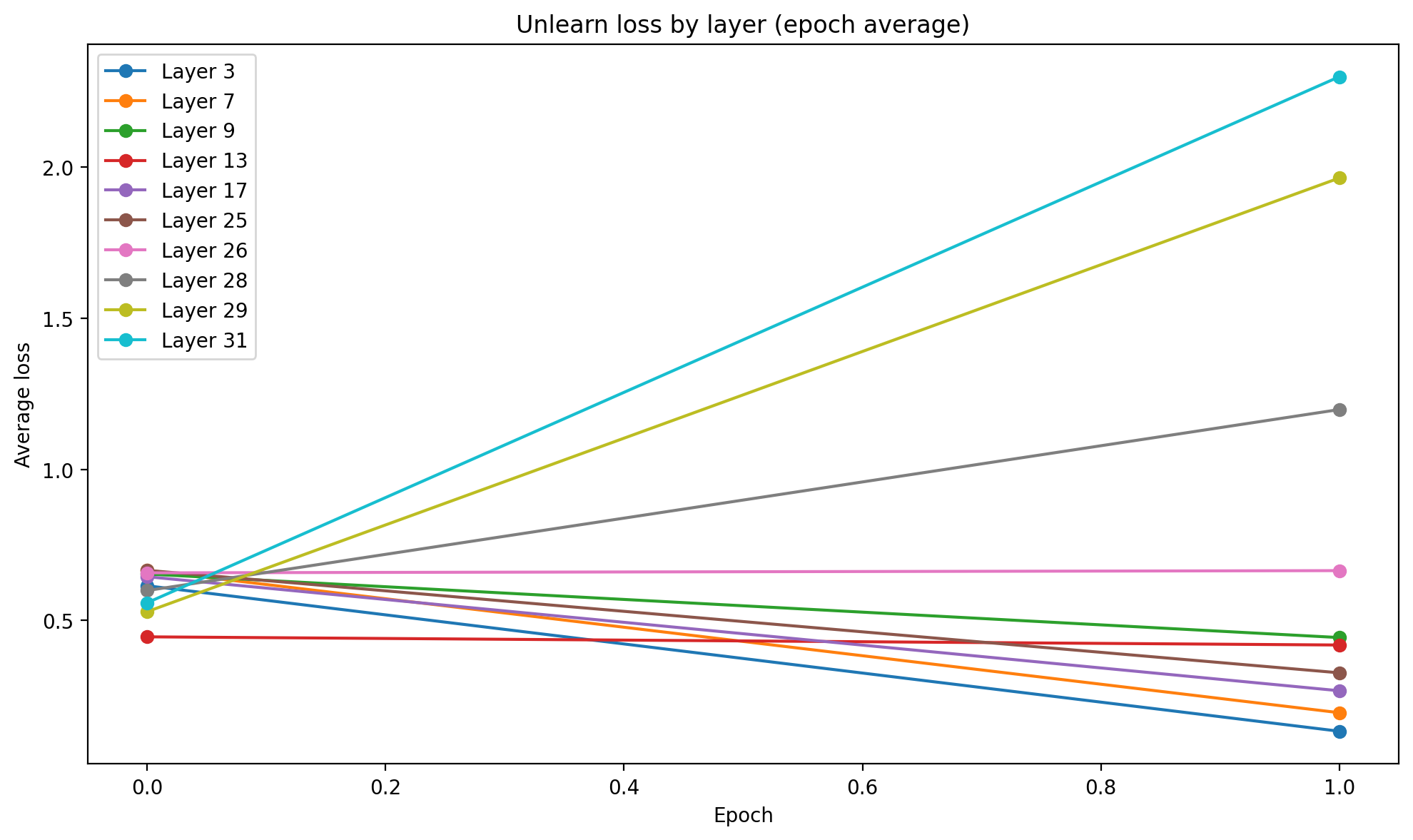}
    \caption{Olmo3-7B}
\end{subfigure}

\caption{Toxicity unlearning loss curves for each model.}
\label{fig:tox_unlearn_loss_curves}
\end{figure*}

For toxicity unlearning, we use the toxic split of ParaDetox \cite{logacheva-etal-2022-paradetox} as forget data. To train the logistic-regression probes, we use the human-annotated training split of TRuST \cite{atil2026justliketrust}. We again use WikiText as retain data and evaluate retained utility with MMLU. To evaluate toxicity, we use the challenging subset of RealToxicityPrompts \cite{gehman2020realtoxicityprompts} and score outputs 
with the BERT-based TRuST classifier \cite{atil2026justliketrust}, which we chose because it generalizes better than alternative toxicity classifiers. For our main toxicity experiments, we update nine layers, selecting three layers each from early, middle, and late depth regions. We analyze alternative layer choices in Section \ref{sec:layers_toxicity}. The final $\alpha$ values are 43, 78, 5.7, and 0.41 for Llama, Mistral, Qwen, and Olmo respectively. 

Given our limited compute budget, we focus on four widely used, smaller, open-source 7--8B models: Llama3.1-8B \cite{grattafiori2024llama3herdmodels}, Mistral-7B \cite{jiang2023mistral7b}, Olmo3-7B \cite{olmo2025olmo3}, and Qwen2.5-7B \cite{qwen2025qwen25technicalreport}.

\section{Results}
We first analyze the learned toxicity probes to better understand the geometry of toxicity representations. We then evaluate our methods on dangerous-knowledge and toxicity unlearning, examine where toxicity unlearning occurs, 
and conclude with an analysis of how many and which layers work best. 




\begin{figure}[t!]
    \centering
    \includegraphics[width=0.47\textwidth]{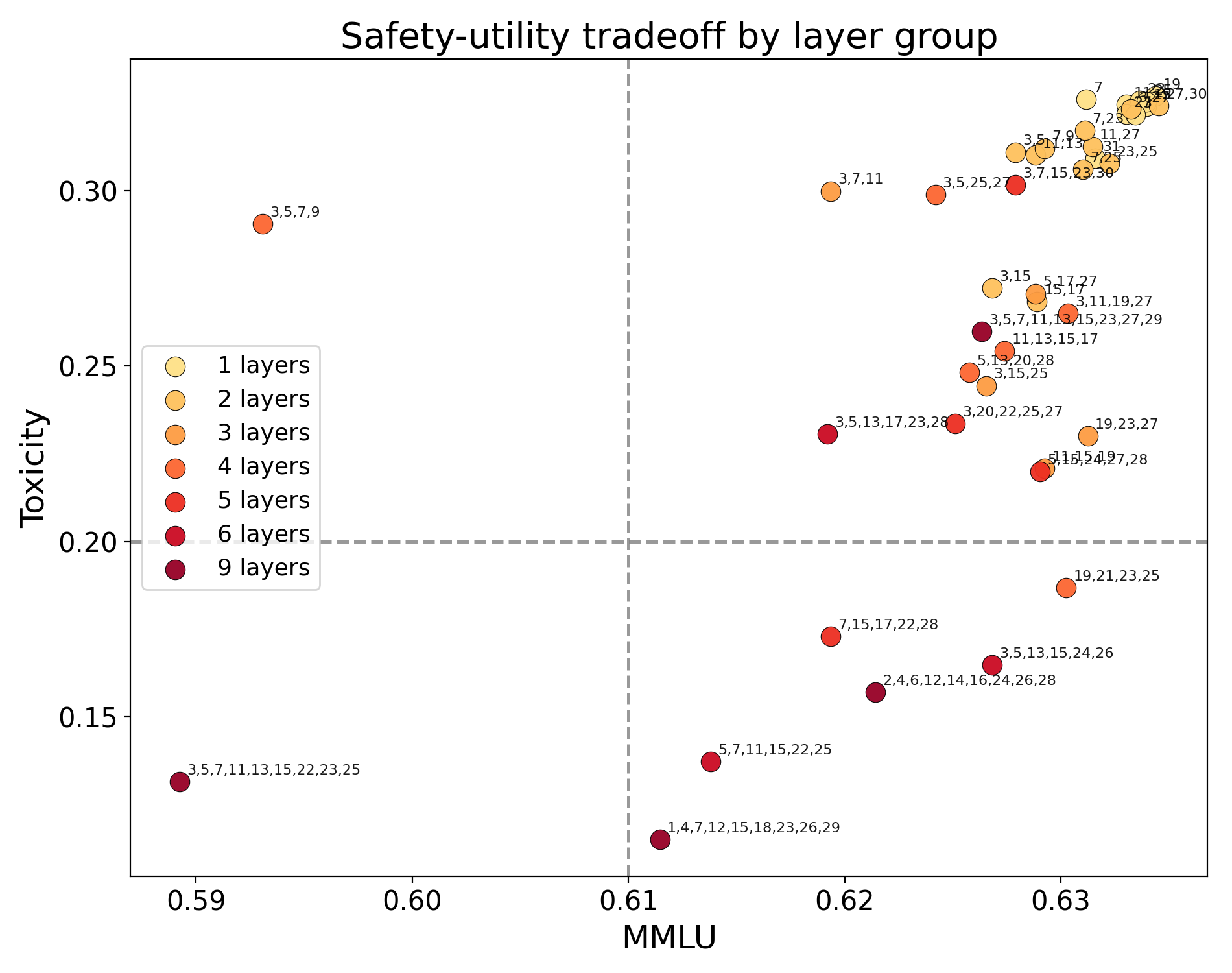}
    \caption{\small The effect of number and identity of the layers on toxicity unlearning and general capability}
    \label{fig:layer_analysis_toxicity}
\end{figure}

\subsection{Toxicity Probe Analysis}
\label{sec:probe_analysis}

To better understand the internal geometry of toxicity representations and obtain principled unlearning directions, we train logistic regression probes at multiple layers of Llama-3.1-8B and analyze the pairwise cosine similarity between the learned weight vectors. Figure \ref{fig:probe_similarity} shows that toxicity directions are largely layer-specific: while adjacent layers share moderate similarity, most cross-layer pairs exhibit cosine similarities below 0.5, especially in the early layers. This indicates that the internal representation of toxicity rotates substantially as information propagates through the network. 
In the late layers, however,
similarities are higher, suggesting a somewhat more consistent toxicity representation near the output. Overall, these results indicate that a single shared direction is insufficient to characterize toxicity across the full depth of the model, motivating our use of layer-specific probes.


Figure \ref{fig:probe_weight_analysis} shows the weight distribution of the probe trained at layer 11 of Llama3.1-8B. The distribution is approximately symmetric and centered close to zero, suggesting that the learned toxicity direction is dense and distributed rather than dominated by a small number of extreme dimensions. The spread of the distribution, spanning roughly $[-0.45,0.45]$, indicates that many dimensions contribute meaningfully to the toxicity direction, with both positive and negative weights. This supports our use of the full probe vector as an unlearning direction and helps explain why sparse or single-neuron interventions are unlikely to be sufficient.

\subsection{Dangerous Knowledge Unlearning}



Figure \ref{fig:danger_un_results} presents the accuracy results for dangerous-knowledge unlearning. As shown, our approach outperforms prior methods for all models except Llama3.1-8B, while preserving retained utility well. All methods remain challenging for Qwen and Olmo models. Table \ref{tab:s_unlearning_resuts} shows our method achieves the best S-unlearning tradeoff on all models except Llama3.1-8B.


\subsection{Toxicity Unlearning}
Figure \ref{fig:tox_unlearn_results} presents the accuracy results for toxicity unlearning. As shown, our method yields the largest 
reductions in toxicity balanced by limited reduction in general knowledge. 
In contrast, RMU and AdapRMU only slightly reduce toxicity. 
As shown in Table \ref{tab:s_unlearning_resuts}, our method also has the highest S-unlearning scores.
The magnitude of improvement varies by model, with the strongest gains on Mistral and Llama, but our method consistently reduces toxic generation.

\subsection{Where Does Toxicity Unlearning Occur?}
To understand which layers are most affected during unlearning, we plot the layer-wise unlearning loss throughout training in Figure \ref{fig:tox_unlearn_loss_curves}. Early layers begin with relatively low loss and remain low or decrease further, suggesting that they encode less discriminative toxicity signal. Middle layers exhibit moderate initial losses and more mixed dynamics. Late layers show the strongest dynamics, typically starting with the highest losses and changing the most during training, consistent with the stronger toxicity representations suggested by our probe analysis in Section \ref{sec:probe_analysis}.

These broad trends hold across models, but the trajectories are model-specific. Mistral-7B (Figure \ref{fig:tox_unlearn_loss_curves}a) shows the most consistent decrease across layers. In Olmo3-7B (Figure \ref{fig:tox_unlearn_loss_curves}d), 
late layers increase sharply, but other layers decrease. In Llama3.1-8B (Figure \ref{fig:tox_unlearn_loss_curves}b), 
earlier layers increase, but mid and late layers decrease. Qwen2.5-7B (Figure \ref{fig:tox_unlearn_loss_curves}c)  exhibits the strongest divergence, with some late layers reaching loss values above 80, indicating that 
here the representations resist orthogonalization more strongly, possibly due to architectural differences in how toxicity is distributed. 
Overall, these results suggest that the geometry of toxicity is architecture-dependent and that the best layer selection strategy may vary across models.

\subsection{Layer Ablation Study for Toxicity}
\label{sec:layers_toxicity}

We analyze both the \emph{number} and the \emph{identity} of the updated layers using Llama3.1-8B, the model where our method performs best. 
Figure \ref{fig:toxicity_vs_numlayers} in the Appendix shows that increasing the number of updated layers generally improves unlearning, although not monotonically: one- and two-layer interventions 
stay near the baseline, while groups of three or more layers yield larger reductions on average. 
Variation across layer groups of the same size further shows that \emph{which} layers are updated matters nearly as much as \emph{how many}. This is consistent with prior work showing that 
changes in transformer representations 
across depths 
are not monotonic \citep{liu-etal-2019-linguistic,liu-etal-2024-fantastic}.

Figure \ref{fig:layer_analysis_toxicity}, plotting toxicity by accuracy in MMLU for different combinations of unlearning layers, shows that layer identity strongly affects safety-utility tradeoff and unlearning. Single-layer interventions, and many two-layer groups, remain in the high-toxicity region, whereas several larger groups reduce toxicity substantially, especially those involving middle and later layers. 
The large variation among groups of the same size further suggests non-additive layer interactions. 
Overall, results suggest that effective toxicity unlearning benefits from coordinated updates across multiple layers, with middle-to-late layers serving as especially useful targets. Further, there is room for large reduction in toxicity with effective choice of layers with little change in general knowledge capabilities.

\section{Conclusion}
\vspace{-0.1cm}
We argued that unlearning in LLMs should be treated as a goal-dependent problem rather than as a single generic intervention. Motivated by mechanistic differences between factual knowledge and toxicity, we introduced a cosine-based, meta-learned variant of RMU for dangerous-knowledge unlearning and then showed that this same approach does not transfer well to toxicity. For toxicity, we instead proposed a multi-layer objective based on layer-specific probe directions.

Across four open-source 7-8B models, our methods performed strongly for both dangerous knowledge and toxicity. Our proposed S-unlearning summary metric makes it easier to compare methods, and to capture the tradeoffs in unlearning.
We presented analyses showing that toxicity directions vary across layers and that effective toxicity unlearning benefits from coordinated updates across multiple depths. 
Finally, our comparison of unlearning for two types of language function that have distinct mechanisms within LLMs suggests that unlearning should be approached as a family of methods that require customization to specific language functions. 

\section*{Limitations}
We focus on two types of unlearning goals in this work, but there are others as well. Further, we focus only on 4 small LLMs because of a limited computational budget.
\bibliography{references, anthology}

\appendix
\newpage

\section{Figures}
\begin{figure}[th!]
    \centering
    \includegraphics[width=0.49\textwidth]{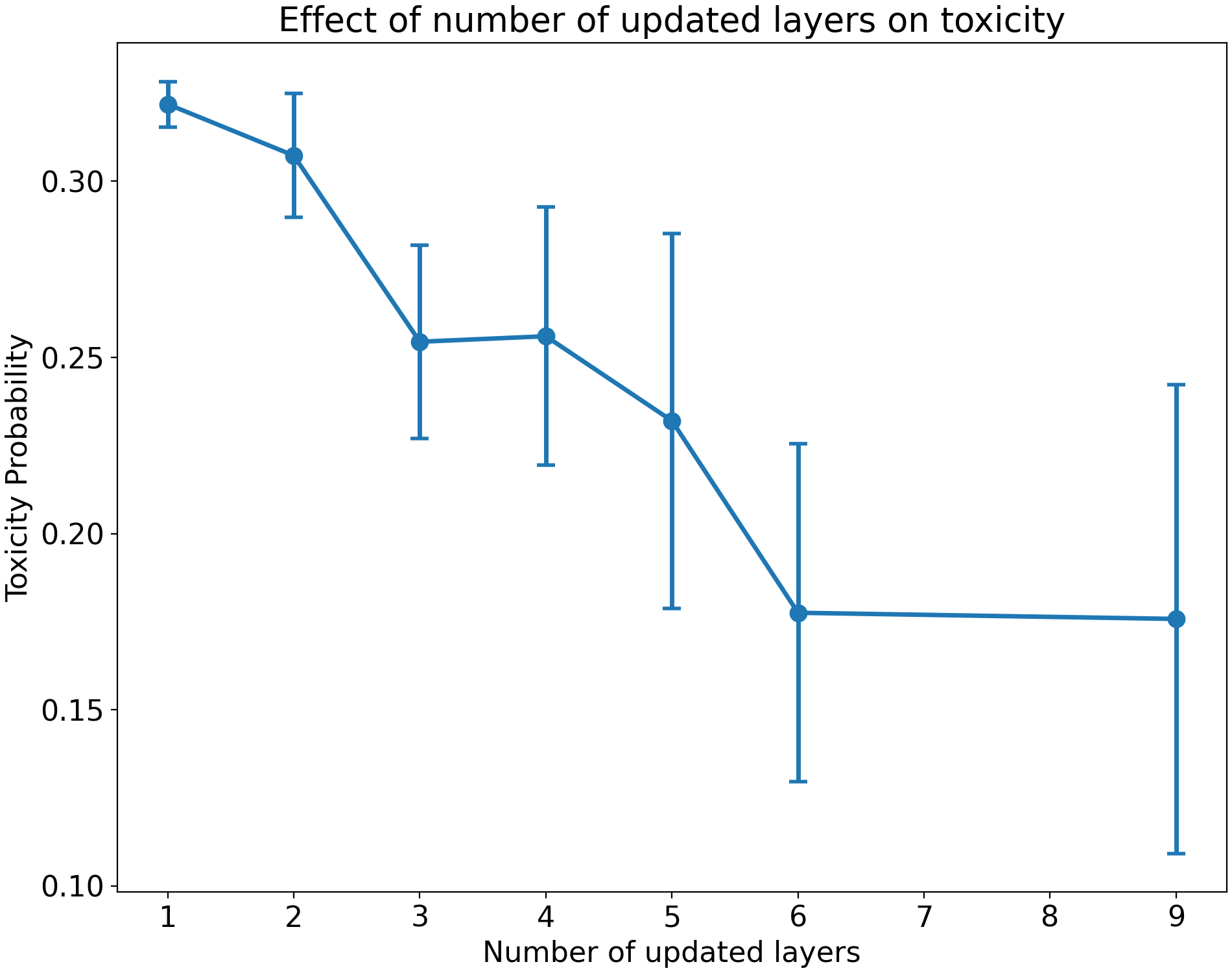}
    \caption{\small The effect of number of layers on toxicity unlearning}
    \label{fig:toxicity_vs_numlayers}
\end{figure}

\newpage
\section{Hyperparameters and Computing Infrastructure}
We have finetuned the models for two epochs on up to 4 Nvidia RTX A6000. Each run took about an hour. For the reinforcement learning, we used a learning rate of 1e-2, and for the main finetuning, we used 5e-5. We used Adam optimizer for all experiments.
\bibstyle{acl_natbib}

\end{document}